%% file: main.tex
 \documentclass[sigconf,nonacm]{acmart}

\AtBeginDocument{%
  }

\usepackage{graphicx}
\usepackage{booktabs} 
\usepackage{multirow} 
\usepackage{siunitx}  
\usepackage{array}    
\usepackage{caption}  
\usepackage{enumitem} 
\usepackage{geometry} 
\definecolor{darkgreen}{rgb}{0,0.5,0}
\begin{document}

\title[Knowledge Graphs and LLM Reasoning for Warehouse Planning Assistance]{Leveraging Knowledge Graphs and LLM Reasoning to Identify Operational Bottlenecks for Warehouse Planning Assistance}
\subtitle{A PREPRINT}

\author{Rishi Parekh}
\affiliation{%
  \institution{Quantiphi}
  \country{Mumbai,India}}
 \email{rishi.parekh@quantiphi.com}

\author{Saisubramaniam Gopalakrishnan}
\affiliation{%
  \institution{Quantiphi}
  \country{Bengaluru,India}}
  \email{saisubramaniam.gopalakrishnan@quantiphi.com}

\author{Zishan Ahmad}
\affiliation{%
  \institution{Quantiphi}
  \country{Bengaluru,India}}
  \email{zishan.ahmad@quantiphi.com}

\author{Anirudh Deodhar}
\authornote{Corresponding author}
\affiliation{%
 \institution{Quantiphi}
 \country{Mumbai,India}}
\email{anirudh.deodhar@quantiphi.com}




\renewcommand{\shortauthors}{A PREPRINT}
\begin{abstract}
Analyzing large, complex output datasets from Discrete Event Simulations (DES) of warehouse operations to identify bottlenecks and inefficiencies is a critical yet challenging task, often demanding significant manual effort or specialized analytical tools. We propose a novel framework that addresses this challenge through a powerful integration of Knowledge Graphs (KGs) and Large Language Model (LLM)-based agents. Our framework first transforms raw DES output data into a semantically rich KG, which uniquely captures the intricate relationships between simulation events and entities (e.g., suppliers, packages, workers, equipment), overcoming unstructured log limitations for robust analysis.
On this KG, our novel LLM-based agent employs a sophisticated iterative reasoning mechanism that interprets complex natural language questions by generating insightful, interdependent sub-questions sequentially. Each sub-question is formulated one at a time, crucially conditioned on the evidence from answers to previous ones. For each individual sub-question, a multi-step process then generates precise Cypher queries for KG interaction, extracts relevant information, and performs crucial self-reflection to identify and correct potential errors. This adaptive, iterative, and self-correcting reasoning progressively pinpoints operational issues and diagnoses root causes, mimicking human investigative analysis.
We evaluate our approach using an example warehouse DES setup, systematically introducing typical bottlenecks like equipment breakdowns and supplier arrival irregularities. For operational questions, our proposed pipeline using step-wise thinking demonstrates significantly higher pass rates compared to traditional baseline methods, achieving near-perfect performance in identifying key inefficiencies. Furthermore, for more complex investigative questions, we qualitatively showcase our framework's superior diagnostic capabilities through three case studies, highlighting its proficiency in uncovering subtle and interconnected inefficiencies often missed by traditional methods.
This work attempts to bridge the gap between simulation modeling and advanced AI-driven data analysis informed by the broader advancements in KG+LLM, offering a more intuitive and potent method for extracting actionable insights from simulation outputs, thereby dramatically reducing time-to-insight and paving the way for automated, intelligent warehouse inefficiency evaluation and diagnosis for industrial data analysis. 

\end{abstract}

\maketitle

\section{Introduction}
Modern warehouses are complex systems, defined by intricate interactions between resources (personnel, equipment), processes (receiving, storage, picking, packing, shipping), and physical layouts \cite{lee2018design, de2007design, gu2007research}. Discrete Event Simulation (DES) has emerged as a powerful technique for modeling these systems \cite{banks2005discrete, law2000simulation}, allowing stakeholders to evaluate performance, test design alternatives, and understand system dynamics before implementation. Despite its power, analysis of the voluminous and highly granular output data generated restricts the full exploitation of DES.
Typical DES runs produce extensive event logs, time-series data on resource states, and detailed queue statistics, capturing micro-level system behavior. Transforming this raw data into actionable intelligence—such as precisely identifying performance bottlenecks, diagnosing root causes of delays, or pinpointing underutilized resources—presents a non-trivial analytical hurdle. Conventional approaches, frequently reliant on manual inspection of aggregate statistics \cite{law2000simulation} or development of custom scripts tailored to specific simulation outputs, are not only time-intensive and error-prone but, critically, often fail to uncover complex, emergent system behaviors and hidden inefficiencies arising from the interplay of numerous components.

The need for more sophisticated and efficient analysis methods is further amplified by the increasing integration of simulation with real-time operational data, particularly within the paradigm of Digital Twins (DTs) \cite{AGALIANOS20201636, leng2019digital, rasheed2020digital}. While DTs aim to provide a synchronized virtual counterpart to physical systems for monitoring and decision-making, the fundamental challenge of interpreting the state and behavior of the simulation or DT persists. Whether analyzing historical simulation runs or near real-time DT states, extracting clear insights remains a bottleneck in itself. AI is playing an increasingly important role in all fields; warehouse logistics is no exception \cite{app13116746, min2010artificial, ivanov2019impact, zhong2017intelligent}. However, to effectively address the specific analytical bottleneck in DES and DT data interpretation for warehouses, there is a pressing need for advanced AI-driven approaches capable of unlocking the rich insights embedded within this complex data, thereby truly enhancing warehouse planning and operational control.

To address these limitations of conventional DES data analysis, our work proposes a novel framework integrating Knowledge Graphs (KGs) \cite{hogan2021knowledge} and Large Language Model (LLM)s \cite{zhao2023survey, pan2023large} for bottleneck identification through natural language queries, leading towards intelligent warehouse planning assistance. The core idea of our work, supported by research on domain-specific and event KGs \cite{abu2021domain}, is to first structure the complex relational data generated by DES using a KG.
Representing simulation output as a graph allows the intricate dependencies and flows within the warehouse system to be explicitly captured and queried. While KGs are increasingly applied to analyze real-world industrial and supply chain data for enhanced visibility and risk management \cite{noy2019industry, kosasih2024towards}, their application to the specific domain of simulation output data remains relatively unexplored.
This work leverages KG technology to achieve deeper simulation understanding, thereby supporting strategic and operational warehouse planning.

Building upon the structured representation provided by the KG, our framework employs LLM-based agents \cite{guo2024large, react22} to enable intuitive interaction with the simulation data \cite{xia2024llm}, directly aiding warehouse planners. LLMs possess powerful natural language understanding and generation capabilities, allowing users—such as operations analysts or industrial engineers involved in warehouse design, optimization, and day-to-day planning, who may lack deep expertise in graph query languages—to pose questions in natural language.
The LLM agent in our framework is not merely an intermediary but is designed with an iterative reasoning \cite{cot22, luo2023reasoning} mechanism for the in-depth diagnostic analysis crucial for warehouse planning. When presented with a complex natural language query regarding warehouse performance or a planning scenario, the agent autonomously generates a sequence of insightful sub-questions. Each sub-question is formulated one at a time, strategically conditioned on the evidence and insights gathered from answers to previous sub-questions directed at the KG. For each such sub-question, the agent then generates precise NL-to-Graph Cypher queries for KG interaction \cite{hornsteiner2024real, mandilara2025decoding}; retrieves relevant information; and performs crucial self-reflection \cite{huang2022large, madaan2023self} to validate its findings and correct potential errors in its analytical pathway. This translation to Cypher, instead of SQL, is a deliberate choice to leverage the KG's native structure; it allows for more expressive queries on complex operational patterns while avoiding the cumbersome joins typical of SQL on graph-like data, a distinction supported by recent text-to-query benchmarks \cite{sivasubramaniam2024sm3}. This synthesis involves not just presenting raw data but interpreting patterns, identifying anomalies (like bottlenecks within specific warehouse zones or affecting critical operational sequences), and inferring potential root causes based on the relationships and event sequences captured within the KG. This approach moves significantly beyond simple data retrieval towards AI-driven analysis and explanation, vital for informed warehouse planning and decision-making.

This synergistic integration of Knowledge Graphs (KGs) and a reasoning LLM-agent transforms a warehouse Digital Twin (DT) from a predominantly passive simulation environment into an interactive, explainable knowledge base and an intelligent assistant for warehouse planners. Consequently, planners can interact with this enhanced DT using natural language to probe multifaceted operational scenarios, diagnose underlying causes of inefficiencies and complex performance deviations, understand the impact of variability (e.g., in supplier arrivals or equipment uptime), evaluate alternative operational strategies more deeply, and proactively identify potential bottlenecks in proposed warehouse layouts or future operational plans—all without the need to manually decipher voluminous simulation logs or write complex scripts. This heightened transparency and AI-driven decision support significantly improves decision-making agility and the strategic depth of warehouse planning. Our work thus provides a robust and novel methodology to bridge advanced AI with established simulation techniques, paving the way for more effective and intelligent warehouse operational management.

The main contributions of this paper are:
\begin{itemize}
    \item \textbf{Novel Supply Chain Application for Bottleneck Identification from DES:} To the best of our knowledge, this work presents the first application combining Knowledge Graphs (KGs) and Large Language Model (LLM) agents specifically for analyzing output data from Discrete Event Simulations (DES) of warehouse operations to identify bottlenecks and inefficiencies.
    \item \textbf{Bridging Simulation and Generative AI:} We establish a methodological bridge between traditional DES analysis techniques and modern AI capabilities offered by KGs and LLMs, proposing a more powerful and intuitive paradigm for interpreting simulation data.
     \item \textbf{Framework Design:} We detail a comprehensive framework encompassing the ontological construction of a KG from DES output data, and the design of an LLM-agent equipped with a novel iterative reasoning mechanism (featuring sequential, conditioned sub-questioning, Cypher generation for KG interaction, and self-reflection) for effective operational performance analysis and bottleneck diagnosis.
    \item \textbf{Experimental Validation Plan:} We propose and evaluate experiments focused on warehouse simulation scenarios with datasets comprising both operational and investigative questions, designed to validate the effectiveness of our KG+LLM framework in identifying operational bottlenecks and enhancing analytical efficiency compared to established baseline methods.
\end{itemize}

\section{Related Work}
\subsection{Discrete Event Simulation and Knowledge Graphs in Warehouse Operations}
Discrete Event Simulation (DES) is extensively employed in logistics and warehousing to model and analyze diverse operational facets.
The outputs from such DES models typically furnish key performance indicators (KPIs) like overall system throughput, queue lengths at different processing stages, waiting times for entities (e.g., orders, products), and the utilization rates of critical resources \cite{banks2005discrete, law2000simulation}. In the contemporary Logistics 4.0 landscape, DES is also increasingly recognized as a fundamental component of Digital Twins (DTs). In this role, DES can function as the \textit{cyber twin}, potentially updated with real-time operational data to mirror physical system states and behaviors \cite{rasheed2020digital, leng2019digital, AGALIANOS20201636}.
A primary analytical objective when working with simulation outputs is the identification of performance bottlenecks. While traditional statistical indicators including average queue lengths, waiting times for entities, and resource utilization
rates \cite{banks2005discrete, law2000simulation} are valuable for initial assessments and identifying obvious areas of concern, they often provide only surface-level insights.
Such methods may not adequately reveal the underlying root causes of identified bottlenecks, particularly those that emerge from complex interactions between multiple system components or from dynamic and fluctuating operational conditions. Consequently, there is a clear and persistent need for analytical methods that can transcend simple statistical thresholds to offer deeper diagnostic capabilities, thereby enabling a more comprehensive understanding of system inefficiencies and their origins.

Artificial intelligence (AI) is increasingly being leveraged to optimize various facets of warehouse operations, including automation and decision support \cite{app13116746, sodiya2024ai, min2010artificial, ivanov2019impact}. Within this trend, Knowledge Graphs (KGs) have emerged as a powerful technology for representing and reasoning over complex, interconnected data in industrial domains \cite{noy2019industry}. KGs are increasingly utilized for diverse applications such as enhancing operational visibility across supply chains, mapping intricate supplier networks, tracking materials and products, managing operational and supply chain risks, optimizing inventory levels, ensuring product traceability, and monitoring sustainability initiatives \cite{saidi2025modeling}. For instance, KGs have been developed to improve robot operations in warehouses \cite{kattepur2019roboplanner} and to create digital twin-enabled dynamic spatial-temporal knowledge graphs for optimizing resource allocation in production logistics \cite{zhao2022digital}.

While the application of KGs to real-world industrial data spanning supply chains, manufacturing processes, and asset management is rapidly advancing and demonstrating significant value, our review indicates a noticeable gap in the application of KG technology specifically to structure, analyze, and interpret the output data generated from simulations, such as DES. This type of data is critical for effective operational planning and strategic decision-making. Much of the existing work on KGs in industrial contexts focuses on modeling the physical system itself, its components, or real-time data streams from sensors and IoT devices. The opportunity to transform the rich, relational event data and temporal sequences produced by detailed simulation runs into semantically rich KGs for in-depth performance analysis and bottleneck diagnosis remains largely untapped. This represents a significant opportunity to leverage the structural and semantic strengths of KGs for achieving a deeper understanding of simulated systems, thereby enhancing simulation-driven planning and optimization.

\subsection{Integrating LLMs and Knowledge Graphs for Industrial Data Analysis}
The advent of Large Language Models (LLMs) \cite{zhao2023survey} has introduced transformative capabilities in natural language understanding, generation, and reasoning. The integration of LLMs with KGs is increasingly recognized as a powerful combination  \cite{pan2023large}, creating a synergy that aims to develop AI systems that are both deeply knowledgeable and intuitively conversational. In this paradigm, KGs serve to ground LLMs with factual, structured knowledge, which can help mitigate issues like hallucinations and improve the accuracy and reliability of LLM-generated responses \cite{agrawal2023can}. Conversely, LLMs can make the rich information stored in KGs more accessible to a wider range of users by enabling natural language querying and interaction, abstracting away the need for specialized query languages \cite{zou2024q2cypher}. 

Several patterns for integrating KGs and LLMs have emerged. One common approach is KG-enhanced LLMs, where KGs are leveraged either during the LLM's pre-training or, more frequently, at inference time; Retrieval-Augmented Generation (RAG) is a prominent technique in this category, using KGs or other external sources to inform and contextualize LLM generation \cite{muneeswaran2024mitigating}. Conversely, LLM-augmented KGs employ LLMs to assist in various stages of the KG lifecycle, including construction from unstructured text, knowledge base completion (like link prediction or entity resolution), enrichment of KG embeddings, or generating textual descriptions from graph data (KG-to-text) \cite{pan2023large}. A third pattern involves synergized LLMs + KGs, characterized by a deeper, often bidirectional integration, frequently featuring LLM-based agents that can reason over, interact with, and manipulate KGs to perform complex, multi-step tasks \cite{jiang2024kg, luo2023reasoning}; the proposed framework aligns with this synergistic approach.

\subsection{KG-LLM Systems in Industrial Applications and the Identified Gap for Simulation Analysis}
The synergistic combination of KGs and LLMs is beginning to find applications in various industrial scenarios. For example, researchers have explored integrating KGs and LLMs for enhanced querying in industrial environments \cite{hovcevar2024integrating}, using LLMs and context-aware prompting to improve access to manufacturing knowledge \cite{monkaaenhancing}, and developing LLM-based assistants for warehouse operations as part of broader AI-driven logistics optimization efforts \cite{ieva2025enhancing}. Furthermore, KG-enhanced LLMs have been proposed for domain-specific question answering systems in technical fields \cite{li2024knowledge}.

However, despite these advancements, the application of KG-LLM systems to the unique challenges of analyzing DES output data for operational insights, particularly for complex tasks like iterative bottleneck diagnosis and root cause analysis, remains largely unexplored. While frameworks like SparqLLM \cite{arazzi2025augmented} have investigated the use of RAG and query templates to improve the reliability of LLM interactions with KGs in industrial settings, the specific application of such LLM+KG agent systems to analyze the unique structure and temporal nature of DES \textit{output data} for operational insights (like bottleneck diagnosis) is largely unexplored. There is a critical need to explore how effectively LLM-based agents, equipped with reasoning capabilities, can:
\begin{itemize}
\item Transform complex natural language questions about  DES output simulated warehouse performance into precise, executable queries over a KG.
\item Iteratively refine their understanding and analytical path based on the evidence retrieved from the KG.
\item Synthesize information from disparate parts of the KG to diagnose operational issues and explain their findings in an intelligible manner.
\end{itemize}

The reliability of LLM-driven query generation, the efficacy of iterative reasoning over simulation-specific KGs, and the overall ability of such integrated systems to provide actionable, explainable insights for DES-based warehouse planning and analysis constitute open research areas that this work aims to address. Our framework specifically focuses on bridging this gap by proposing a novel LLM-based agent that employs an iterative, self-correcting reasoning process over KGs derived from DES outputs to automate and enhance the identification and diagnosis of warehouse inefficiencies.

\section{Discrete Event Simulation -- Scenario Design of Warehouse Discharge}
This study is based on the data generated by an in-house discrete-event simulation (DES) model that includes operations of a warehouse facility engaged in the unloading, internal transport, and storage of incoming packages. The simulation is designed to replicate real-world warehouse logistics, capturing the interactions between key resources such as suppliers (trucks), workers, automated guided vehicles (AGVs), forklifts, and storage infrastructure. Key highlights are described here, more information can be found in appendix. 


\subsection{Scenario Configuration}
The following sections outline the detailed scenario configuration used in the simulation model. These parameters and design decisions collectively contribute to the fidelity of the simulation to replicate real-world warehouse operations:
\begin{figure}[!t]
  \centering
  \includegraphics[width=0.99\linewidth]{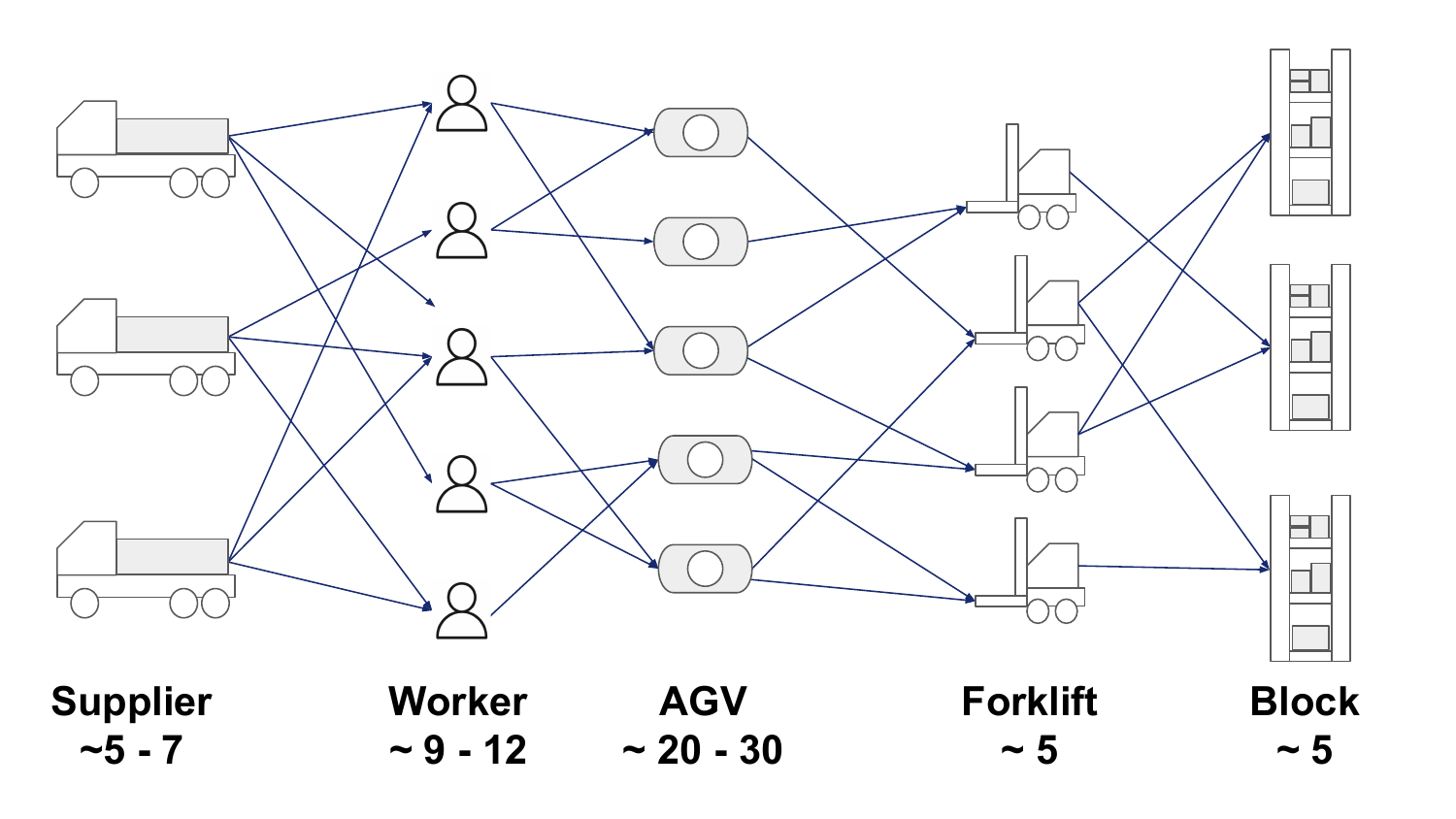}
  \caption{Workflow diagram of the warehouse unloading process, detailing the sequence from supplier deliveries and worker handling, through stages involving Automated Guided Vehicles (AGVs) and forklifts, to packages formed for storage. Numerical figures are for representational purposes only, the exact numbers are provided in text.}
  \label{fig:simulation_flow}
\end{figure}

\subsubsection{Equipment and Resource Specifications:} 
\begin{itemize}
    \item \textbf{Suppliers:} These are external trucks that bring in the packages to the warehouse. Each supplier holds a total of N packages where N is sampled from a distribution ranging from 30 to 35. They move at a speed of 20 km/hr. A total of five suppliers are expected to arrive in simulation. 
    \item \textbf{Workers:} These are 12 employees working at the warehouse. They can move one package at a time at a speed of 2 km/hr.
    \item \textbf{Automated Guided Vehicles (AGVs):} They are automated transporters that carry packages from the worker to the forklift. They can be programmed to traverse pre-specified paths. There are 20 AGVs present in the warehouse. They move with a speed of 3.5 km/hr. The Time taken for the AGV to transport a package depends on the distance it travels which is 140 meters on an average. 
    \item \textbf{Forklifts:} They are man operated machines that can move packages both horizontally and vertically. They pick up packages from AGVs and place them on a given shelf in the storage block. A forklift can move at speed of 5 km/hr. 
\end{itemize}

\subsubsection{Process Flow:} 
The unloading and storage process follows a structured flow:

\begin{enumerate}
    \item \textbf{Supplier Arrival}: On arrival, a supplier truck heads to the parking area and waits to be assigned to an unloading dock. Once an unloading dock becomes available, the supplier moves towards it and starts unloading.  

    \item \textbf{Unloading Operation}: Each supplier is assigned a team of four workers. Upon reaching an available unload spot, workers begin transferring packages from the supplier to a pre-defined waiting point.

    \item \textbf{Worker to AGV Handoff}: Workers wait for the arrival of an AGV at the waiting point. Upon arrival, the package is loaded onto the AGV and the worker returns to the supplier to repeat the process. The worker handling time is determined by the distance between the supplier and the waiting point, adjusted for the walking speed.

    \item \textbf{AGV Transit}: The AGV transports the package to the appropriate block-specific pickup point, with travel time determined by distance and AGV speed.

    \item \textbf{Forklift Transfer}: A dedicated forklift on the block collects the package from the AGV and stores it in one of the available bays. The forklift operation time includes the travel time and a stochastic storage time drawn from a distribution ranging between 60 to 90 seconds. The storage space has a cluster of 15 bays with each bay having 3 shelves. Each storage block can store 45 packages. There are five such blocks in the warehouse, making the total capacity 225.
    
\end{enumerate}


\subsubsection{Operational Assumptions:} 
\begin{itemize}
\item Suppliers arrive at regular intervals of 30 minutes, subject to yard capacity constraints (maximum of 3 unloading simultaneously).
\item Each worker team of four operates exclusively with its assigned supplier and unload dock during the unloading process.
\item AGVs are dynamically dispatched to waiting points and are assigned to packages in a First-In-First-Out (FIFO) manner.
\item Forklifts serve incoming packages at the pickup point using a FIFO approach and are restricted to their respective blocks.
\item Package handling times for workers and AGVs are determined by distance and speed; storage time for forklifts includes a stochastic component.
\item Storage allocation for a block is stochastic.
\end{itemize}

\subsubsection{Data Extracted from Simulation} Following data is captured from each simulation run
\begin{itemize}
    \item \textbf{Process and equipment specific:} The process specific data, including the equipment ID, arrival time, process initiation time, waiting time, process completion time. This includes suppliers, workers, AGVs and forklifts. 
    \item \textbf{Package specific:} For each individual package, we capture a unique package ID and log key timestamps throughout the material handling process. These include the time the package is picked up from the supplier, the waiting time at the transfer point, the time the package is loaded onto and departs with an AGV, the time of arrival at the storage block, and the final timestamp when the package is placed into storage.
\end{itemize}


\subsection{Generation of Evaluation Questions}
Based on data generated from the normal operating scenario, two types of questions were formulated for the evaluation of analytical capabilities:
\subsubsection{Operational Questions:} A set of 25 distinct operational questions (see Table \ref{tab:operational_questions}) was created to assess the proficiency in retrieving specific factual information and performing straightforward analyses using the simulation output. These questions were designed to cover various aspects of the simulated operation, with an approximately uniform distribution across key entities and stages such as supplier interactions, worker activities, AGV and forklift utilization, and package flow.
\subsubsection{Investigative Scenarios and Questions for Bottleneck Identification:} To specifically evaluate the  capabilities in identifying operational bottlenecks, three distinct investigative scenarios were simulated. Each scenario introduced a specific type of inefficiency into the baseline model, mirroring potential real-world disruptions:
\begin{itemize}
    \item \textbf{Scenario 1: Delay in Stage Transfer:} For a particular supplier, a specific process inefficiency was simulated, primarily introducing intermittent delays within the AGV-to-Forklift (AGV-FL) transfer stage, leading to significantly prolonged overall discharge times for their packages.
    \item \textbf{Scenario 2: Supplier-Specific Processing Delay:} For a particular supplier, targeted inefficiencies were simulated, introducing increased handling and suboptimal task allocation within the unloading and package processing stages, leading to significantly prolonged processing times.
    \item \textbf{Scenario 3: Degraded Forklift Performance:} One specific forklift was modeled to operate with reduced efficiency throughout its designated shift, leading to localized congestion and delays in tasks reliant on that particular forklift.
\end{itemize}
For each of these three systematically perturbed scenarios, a unique investigative question was formulated. The objective of each such question was to task the framework with identifying the primary operational bottleneck or pinpointing the most significant performance degradation resulting from the deliberately introduced inefficiency.

\begin{figure}[htbp]
  \centering
  \includegraphics[width=0.9\linewidth]{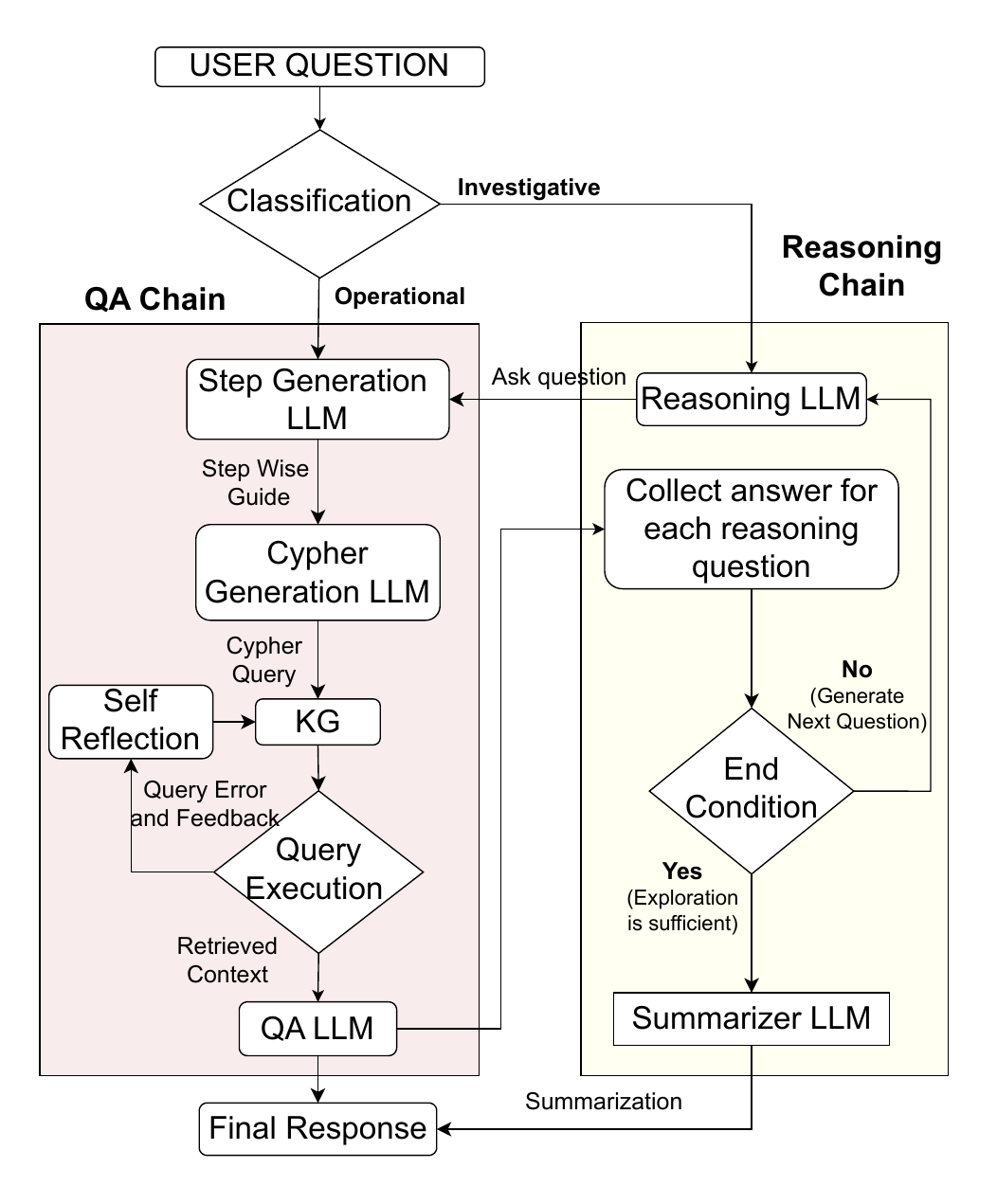}
  \caption{System architecture of the LLM Reasoning Agent comprising two components: the QA Chain and the Reasoning Chain. Queries are first classified as normal or bottleneck. Normal queries follow step-wise guidance for Cypher generation, execution, and self-reflection. Bottleneck queries invoke iterative reasoning, where the agent decomposes the problem into sub-questions, gathers intermediate evidence, and dynamically refines its analysis toward a final answer.}
  \label{fig:qa_architecture_agent}
\end{figure}

\begin{table*}[htbp]
\centering
\caption{Performance on Operational QA by Method and Stage (Pass@k Scores). \\ Direct QA: Single-pass Cypher query generation followed by answer synthesis. SR: Self-Reflection. Step-wise Guide: Question decomposition for structured step generation; each step involves (Cypher query + Answer Generation + Self-Reflection). P@k indicates Pass@k scores.}
\label{tab:operational_qa_results}
\begin{tabular}{@{}c | cc | cc | cc | cc | cc | cc@{}} 
\toprule
\multirow{2}{*}{\textbf{Method}} & \multicolumn{2}{c}{\textbf{Supplier}} & \multicolumn{2}{c}{\textbf{Worker}} & \multicolumn{2}{c}{\textbf{AGV}} & \multicolumn{2}{c}{\textbf{Forklift}} & \multicolumn{2}{c}{\textbf{Package}} & \multicolumn{2}{c}{\textbf{Average}} \\
\cmidrule(lr){2-3} \cmidrule(lr){4-5} \cmidrule(lr){6-7} \cmidrule(lr){8-9} \cmidrule(lr){10-11} \cmidrule(lr){12-13}
 & \textbf{P@1} & \textbf{P@4} & \textbf{P@1} & \textbf{P@4} & \textbf{P@1} & \textbf{P@4} & \textbf{P@1} & \textbf{P@4} & \textbf{P@1} & \textbf{P@4} & \textbf{P@1} & \textbf{P@4}\\
\midrule
Direct QA \textsuperscript{a}         & 0.50 & 0.60 & 0.40 & 0.40 & 0.25 & 0.60 & 0.41 & 0.56 & 0.41 & 0.56 & 0.41 & 0.56 \\
Direct QA + SR \textsuperscript{b}      & \textbf{0.95} & \textbf{1.00 }& 0.55 & 0.80 & 0.60 & 0.60 & 0.73 & 0.80 & 0.73 & 0.80 & 0.73 & 0.80 \\
Step-wise Guide \textsuperscript{c}  & 0.90 & \textbf{1.00} & \textbf{0.74} & \textbf{1.00 }& \textbf{0.83} & \textbf{1.00} & \textbf{0.75} & \textbf{ 1.00} & \textbf{0.90} &\textbf{ 1.00} & \textbf{0.82} &\textbf{ 1.00} \\
\bottomrule
\end{tabular}
\end{table*}
\section{Methodology}
This section briefly outlines the core technical components of the proposed framework. 

\subsection{KG Schema Design for DES Data}
We utilize a custom KG schema tailored to represent the resources (supplier, worker, AGV, forklift, storage) as nodes and movement of each package between resources as edges of the KG. The operational data including timestamps is added as features of these edges and nodes. The KG is constructed from the output logs generated by the DES model through an automated pipeline. See appendix \ref{appendix:KGSchema} for more details.

\subsection{LLM Reasoning Agent}
The LLM reasoning agent utilizes a dual-path architecture, initiated by query classification, for the complex analysis of operational and investigative questions on the simulation output-derived KG. For operational queries, a QA Chain features a Step Generation module that translates the natural language query into structured steps, often breaking down complex questions into simpler sub-queries, each aimed at extracting relevant information via a single, targeted Cypher query. A Cypher Generation module then formulates these formal queries, which are programmatically executed against the KG by a Query Execution and Correction Module that incorporates an error-handling loop. The Answer Synthesis (QA) module subsequently receives and processes the query results (e.g., list of entities, subgraphs, aggregated values), moving beyond mere data presentation to interpret initial patterns and synthesize coherent answers. 

For complex investigative (bottleneck) queries, an Iterative Reasoning Chain is activated: a Reasoning module decomposes the main problem and sequentially generates sub-questions one at a time. Each sub-question then leverages the entire QA Chain—enabling iterative Cypher generation, execution, and focused evidence collection—which allows the agent's overall analytical path to be dynamically refined based on intermediate findings. Upon meeting a sufficiency condition, a Summarizer module performs the final answer synthesis. This stage involves in-depth interpretation of the aggregated graph data to identify patterns indicative of performance issues like bottlenecks 
(e.g., identifying workstations with consistently high incoming flow but low outgoing flow, correlated with long queue times) and potentially suggesting causal factors based on traversing relationships in the KG (e.g., linking a bottleneck to upstream resource unavailability or specific event sequences
thereby delivering a comprehensive diagnostic summary). This architecture synergizes robust, self-correcting query execution with adaptive, iterative reasoning for advanced diagnostics, where each distinct processing module is realized through an independent Large Language Model call, leveraging either a general-purpose foundation model or a smaller, task-specific fine-tuned model. 

\section{Results and Discussion}
We evaluated our LLM agent framework using OpenAI’s GPT-4o (via Langchain QA chains) interacting with a Neo4j knowledge graph through LLM-generated Cypher queries. Interactions were configured with temperature 0.0, top\_p 0.95, and a 4096-token limit. While a zero temperature aims for determinism, minor variability can arise from top\_p sampling or multi-step reasoning dynamics. For operational queries, we employed a QA chain guided by a step-wise approach that decomposed input questions into structured steps—each involving Cypher generation, KG querying, and self-reflection. Performance was measured using the pass@k metric\cite{chen2021evaluating} to assess answer accuracy across 4 attempts. This was benchmarked against two baselines: (i) single-pass Cypher generation with answer synthesis, and (ii) an enhanced version adding post-answer self-reflection. For investigative bottleneck scenarios, our iterative Reasoning chain—refining each step based on accumulated evidence was compared qualitatively against the enhanced baseline and a human expert.  Future work will explore additional specialized reasoning models.

\subsection{Performance on Operational QA}
The experimental results for operational question answering presented in Table \ref{tab:operational_qa_results} highlight the significant advantages of our proposed Guided Iterative Steps approach. While incorporating a self-reflection (SR) mechanism into a direct question-answering pipeline (Baseline: Direct QA + SR) does offer a substantial improvement over a simple, single-pass baseline: Direct QA, our proposed method consistently outperforms both baselines, particularly in achieving comprehensive correctness as indicated by the maximum Pass@4 scores across all operational stages. Qualitative success and failure cases across the approaches are provided in Appendix \ref{operational_qa_success_failure}.

\begin{table*}[htbp]
\centering
\caption{Case Study 1: Why did CamelCargo discharge take longer than usual?}
\label{tab:case_study_camelcargo}
\begin{flushleft}
\footnotesize{\textbf{Note:} The LLM Agent formulates its own sequence of questions based on its reasoning framework and evolving evidence from the KG. Human expert iterative questions for this investigation: (1) What was the difference in discharge time of CamelCargo? (2) What was the average discharge time of other suppliers? (3) How many packages needed to be unloaded by CamelCargo? (4) What was the average time taken per package in terms of worker operation time, AGV arrival time, AGV operation time, and forklift operation time? (5) Calculate derived metrics such as waiting time (AGV end time - forklift start time) (6) Did any package deviate from the average time? If yes, in which category? The LLM generations have been condensed to fit inside the table. Cypher queries shown are highly condensed conceptual representations for brevity.}
\end{flushleft}

\footnotesize 
\begin{tabular}{
    >{\raggedright\arraybackslash}p{0.23\textwidth} 
    >{\raggedright\arraybackslash}p{0.23\textwidth} 
    >{\raggedright\arraybackslash}p{0.23\textwidth} 
    >{\raggedright\arraybackslash\ttfamily}p{0.2\textwidth}  
}
\toprule
\textbf{LLM Agent - Iterative Question} & \textbf{LLM Agent - Plan (Condensed)} & \textbf{KG ResultSet (Condensed)} & \textbf{Cypher Query (Condensed)} \\
\midrule

What is the total unload time for CamelCargo compared to the global average unload time for all suppliers in the Discharge Flow? & 
Calculate CamelCargo's unload duration (discharge\_start to discharge\_end) and compare with global average for all suppliers. & 
CamelCargo Unload Time: 6848s. \newline Global Avg. Unload Time: ~4933.8s. & 
\texttt{MATCH (s:SUPPLIER} \texttt{\{supplier\_id:'CamelCargo'\}) RETURN duration.between(s.discharge\_start, s.discharge\_end); MATCH (s:SUPPLIER) RETURN avg(duration.between(s.discharge\_start, s.discharge\_end))} \\
\cmidrule(lr){1-4}

What is the breakdown of time taken for each stage of the unload process for CamelCargo in the Discharge Flow, and how does it compare to the global average for each stage? & 
Calculate durations for Worker-to-AGV, AGV-to-FL, FL-to-Storage stages for CamelCargo's packages; compare with global averages for these stages. & 
Worker-AGV: ~58s (matches global avg). \newline AGV-FL (CamelCargo): Highly variable (e.g., one 2300s), many > global avg ~422.6s. \newline FL-Storage: Near global avg ~116.4s. & 
\texttt{MATCH (s:SUPPLIER \{id:'CC'\})-[*]->(st:STORAGE) RETURN duration(worker\_stage) AS t\_ws, duration(agv\_stage) AS t\_as, ...; MATCH path RETURNavg(t\_ws),avg(t\_as)...} \\
\cmidrule(lr){1-4}

What are the waiting times for AGVs and forklifts during the CamelCargo unload process in the Discharge Flow, and how do they compare to the global average waiting times for these equipments? & 
Calculate AGV waiting (e.g., worker pick-up end - AGV arrival) and Forklift waiting (e.g., AGV journey end - FL placement start) for CamelCargo; compare to global averages. & 
Avg. AGV Waiting (CamelCargo): Neutral (~12s). \newline Avg. FL Waiting (CamelCargo): Positive (~60.6s, indicating delays). & 
\texttt{MATCH (s:SUPPLIER \{id:'CC'\})-...->(a:AGV) RETURN duration.between(wta.agv\_arrival, wta.worker\_end) AS agv\_w;...avg(agv\_w)...} \\
\cmidrule(lr){1-4}

What is the utilization rate of forklifts during the CamelCargo unload process in the Discharge Flow, and how does it compare to the global average utilization rate for forklifts? & 
Determine FL utilization (total processing time / total active time) for each forklift with CamelCargo packages; compare to global average FL utilization. & 
Individual FL utilization rates for CamelCargo (e.g., FL\_00: ~41.2\%) generally matched their respective global average utilization rates. & 
\texttt{MATCH (a:AGV)-[]->(fl:FL)-[]->... RETURN fl.id, SUM(dur\_proc) / dur\_active AS util; ...avg(util)} \\
\cmidrule(lr){1-4}

What is the utilization rate of AGVs during the CamelCargo unload process in the Discharge Flow, and how does it compare to the global average utilization rate for AGVs? & 
Determine AGV utilization (total processing time / total active time) for each AGV handling CamelCargo packages; compare to global average AGV utilization. & 
AGV utilization rates varied (e.g., AGV\_00: ~38.1\%, AGV\_10: ~55.1\%); some higher/lower than an estimated global average of ~41.4\%. & 
\texttt{MATCH (w:WORKER)-[]->(a:AGV)-[]->... RETURN a.id, SUM(dur\_proc) / dur\_active AS util} \\
% ...
\midrule 
\multicolumn{4}{p{\dimexpr0.23\textwidth + 0.27\textwidth + 0.27\textwidth + 0.19\textwidth + 2\tabcolsep\relax}}{%
\textbf{Agent Summary:} The \textbf{`AGV to FL' stage} for CamelCargo exhibited significant delays (specific instances up to ~2300s versus ~423s average for this stage) and high variability, which was the main contributor to its extended total unload time (6848s vs. ~4934s global average). Forklift waiting times also indicated related delays. Other stages and overall equipment utilization rates were found to be less critical, pointing towards process inefficiencies primarily within the AGV-FL transfer.
} \\
\bottomrule
\end{tabular}
\end{table*}

Our proposed technique takes a different approach than the conventional reliance on a single, monolithic Cypher query for operational questions, an approach prone to brittleness due to its complexity. Instead, it introduces a layer based on question decomposition and structured step-wise guidance generation. Rather than attempting to retrieve and synthesize information in one pass, our agent breaks down each query into a sequence of focused analytical steps. Each step involves targeted Cypher query formulation, execution, and an immediate self-reflection phase to assess and refine the output before proceeding. This step-level interaction enables localized error detection and correction, improving both precision and robustness. This approach contrasts sharply with the baselines, where reflection—if present—occurs only after a full KG interaction, limiting its corrective potential. By embedding reflection within each step, our agent incrementally builds understanding, guided by intermediate results and error signals. This not only reduces query complexity but also strengthens the overall reasoning process. The resulting architecture supports more reliable QA and serves as a foundation for the iterative, evidence-driven reasoning required in investigative tasks. The ability to decompose, validate, and refine sub-questions at each stage enables more accurate KG interactions and enhances the agent's capability in both direct question answering and complex diagnostic scenarios.

\subsection{Qualitative Analysis on Investigative QA}
We present three case studies to evaluate the agent’s effectiveness in handling investigative QA. Due to space constraints, only two are discussed in detail, and the third scenario is provided in Appendix \ref{scenario_3}.

\begin{table*}[htbp]
\centering
\caption{Case Study 2: Why was the discharge slow from 10 to 12.30?}
\label{tab:case_study_timeframe_slowdown}
\begin{flushleft}
\footnotesize{\textbf{Note:} Human expert iterative questions for this investigation: (1) What was the difference in discharge time of suppliers between 10:00 AM and 12:30 PM? (2) What was the average discharge time of other suppliers during this period? (3) What was the average time taken per package across key operations (worker, AGV, forklift) within this timeframe? (4) Calculate derived metrics such as waiting times for equipment or packages. (5) Did any package deviate significantly from average processing times? (6) Do these deviations fall within the 10:00 AM - 12:30 PM window, and which operational categories are primarily responsible? The LLM generations have been condensed to fit inside the table. Cypher queries are highly condensed conceptual representations for brevity.}
\end{flushleft}
\footnotesize 
\begin{tabular}{
    >{\raggedright\arraybackslash}p{0.23\textwidth} 
    >{\raggedright\arraybackslash}p{0.23\textwidth} 
    >{\raggedright\arraybackslash}p{0.23\textwidth} 
    >{\raggedright\arraybackslash\ttfamily}p{0.2\textwidth}  
}
\toprule
\textbf{LLM Agent - Iterative Question} & \textbf{LLM Agent - Plan (Condensed)} & \textbf{KG ResultSet (Condensed)} & \textbf{Cypher Query (Condensed)} \\
\midrule

What was the total unload time for each supplier from 10:00 AM to 12:30 PM, and the global average? & 
Calculate unload duration (discharge\_end - discharge\_start) for each supplier active in the specified timeframe; compute global average. & 
AuroraFarms: 8896s; BlackSheepDist: 6713s; CamelCargo: 5104s. \newline Global Avg. (in timeframe): ~6904.3s. & 
\texttt{MATCH (s:SUPPLIER) WHERE s.discharge\_start/end IN [timeframe] RETURN s.id, duration(...);...avg(duration)} \\
\cmidrule(lr){1-4}

What were the package waiting times for each supplier from 10:00 AM to 12:30 PM, and the global average? & 
Calculate average total processing duration for packages per supplier active in timeframe (summing key stage durations); compute global average. & 
Avg. Pkg. Processing Duration - AuroraFarms: ~760.1s; BlackSheepDist: ~746.9s; CamelCargo: ~564.0s. \newline Global Avg: ~689.9s. & 
\texttt{MATCH (s:SUPPLIER)-[*]->(st:STORAGE) WHERE s.discharge IN [timeframe] RETURN s.id, avg(total\_pkg\_proc\_time);...avg(...)} \\
\cmidrule(lr){1-4}

What was the utilization rate of workers and AGVs for each supplier from 10:00 AM to 12:30 PM, and the global average? & 
Determine worker and AGV utilization (processing time / active time) per supplier in timeframe; compute global averages. & 
AuroraFarms: Varied worker util. (some low, e.g., ~2.6\%), varied AGV util. (some high, e.g., ~86\%). Other suppliers also varied. & 
\texttt{MATCH (s)-[]->(w:WORKER), (s)-[]->(a:AGV) WHERE s.discharge IN [timeframe] RETURN s.id, util(w), util(a); ...avg(...)} \\
\cmidrule(lr){1-4}

What was the supplier waiting time for each supplier from 10:00 AM to 12:30 PM, and the global average? & 
Calculate supplier waiting (discharge\_start - arrival\_time) for suppliers arriving in timeframe; compute global average. & 
Main suppliers (Aurora, BlackSheep, CamelCargo): 0s waiting. \newline DeltaDrops: 52s. Global Avg: ~15.4s. & 
\texttt{MATCH (s:SUPPLIER) WHERE s.arrival\_time IN [timeframe] RETURN s.id, duration.between(s.arrival, s.discharge\_start);...avg(...)} \\
\cmidrule(lr){1-4}

What was the breakdown of time taken for each stage of the package unload process for each supplier from 10:00 AM to 12:30 PM, and the global average? & 
Further dissect stage durations (Worker-AGV, AGV-FL, FL-Storage) for each supplier in timeframe; compare to global stage averages to identify specific inefficiencies per supplier. & 
(Detailed stage breakdown per supplier, e.g., AuroraFarms' AGV-FL stage contributing significantly to its high package processing time.) & 
\texttt{MATCH (s:SUPPLIER)-[stages...]-> (st:STORAGE) WHERE s.discharge IN [timeframe] RETURN s.id,avg(dur\_stage1), avg(dur\_stage2)...} \\
\midrule 
\multicolumn{4}{p{\dimexpr0.23\textwidth + 0.27\textwidth + 0.27\textwidth + 0.19\textwidth + 3\tabcolsep\relax}}{%
    \textbf{Agent Summary:} Operations related to supplier \textbf{`AuroraFarms'} were the main driver. This was evidenced by AuroraFarms significantly longer total unload time (8896s vs. \textasciitilde6904s global average for the period) and higher average package processing/waiting durations (\textasciitilde760s vs. \textasciitilde690s global average). Contributing factors included inefficient worker (some instances as low as \textasciitilde2.6\%) and variable AGV utilization (some instances as high as \textasciitilde86\%) associated with AuroraFarms' packages during this timeframe. Initial supplier waiting times for key suppliers were not a factor.
} \\
\bottomrule
\end{tabular}
\end{table*}
\subsubsection{Scenario 1:} The first study examined why a supplier, \textit{CamelCargo}’s discharge was significantly delayed. The human expert identified a critical symptom: a 38-minute delay at the AGV stage for the final package. When the same question was posed to the baseline method, it broadly attributed delays to varying times at each stage, mentioning the AGV and forklift but failing to isolate or quantify the main bottleneck. In contrast, our LLM-agent (process detailed in Table \ref{tab:case_study_camelcargo}) validated the expert’s observation with precise, data-driven analysis. It first confirmed the overall delay (6,848s vs. a ~4,934s average), then, through sub-questioning, identified the “AGV to FL” transfer as the key issue—highlighting extreme variability and delays. Further analysis of related factors (e.g., normal AGV wait times, localized forklift delays, typical utilization rates) led to a robust conclusion: the bottleneck stemmed from inefficiencies in the AGV-to-FL process. This  demonstrates the agent’s strength in not only aligning with expert intuition but also delivering a more precise and comprehensive diagnosis than the baseline.
\subsubsection{Scenario 2:} The second study addressed a slowdown in warehouse discharge operations observed between 10:00 AM and 12:30 PM. The human expert, analyzing detailed data, noted that AGV operational times seemed longer for most packages between 10:30 AM and 11:11 AM, but was not conclusive whether this was solely due to AGVs or potentially linked to specific workers or forklifts, though other resource timings appeared normal. The baseline approach provided a very general explanation for the overall slowdown in the 10:00 AM - 12:30 PM window, attributing it to the average operational durations of workers (~58s), AGVs (~474s), and forklifts (~118s) without identifying any specific entity or cause for deviation. In contrast, our iterative LLM agent (Table \ref{tab:case_study_timeframe_slowdown}) systematically diagnosed the issue within the given timeframe. It first identified that 'AuroraFarms' had a significantly longer total unload time (8,896s) compared to other suppliers and the period's global average (~6,904s). Subsequent investigation revealed that AuroraFarms, along with 'BlackSheepDist', also exhibited higher average package processing durations. Crucially, the agent pinpointed inefficient worker and AGV utilization linked to AuroraFarms (e.g., some worker utilization as low as ~2.6\% and high AGV utilization peaks suggesting bottlenecks) as key contributing factors, while ruling out initial supplier waiting times. This allowed the agent to determine that the slowdown within the specified timeframe was primarily driven by inefficiencies related to a specific supplier, AuroraFarms, particularly concerning their package processing throughput and associated resource utilization, a far more precise and actionable insight than either the human expert's localized AGV observation or the baseline's generic summary.

\subsection{Discussion on relevance for warehouse planning}
Our framework demonstrates significant relevance as a planning assistant across multiple horizons of warehouse operations. Its high pass@k scores on diverse operational queries (Table \ref{tab:operational_qa_results}) enables planners to obtain precise, real-time visibility into supplier interactions, resource utilization, and package flow—supporting both reliable daily control and agile tactical adjustments. More importantly, the investigative case studies highlight the framework’s ability to move beyond surface-level reporting toward meaningful diagnostic insight. By systematically querying a simulation-derived KG using an LLM-driven reasoning process, the agent effectively isolates root causes of performance issues, revealing subtle bottlenecks and inter-dependencies often missed by traditional analytics. This fusion of DES with GenAI methods offers a more powerful and interpretable warehouse digital twin. As a result, planners are better equipped to make targeted, data-driven interventions—whether through process redesign, resource reallocation, or supplier strategy refinement—ultimately enabling more adaptive, efficient, and informed warehouse planning.

\section{Implications and Limitations}
This work marks a significant step towards automating the intricate analysis of DES outputs, offering warehouse planners a potent tool for rapid diagnostic insights via natural language. 
However, the current study has certain limitations. While our proposed pipeline facilitates KG construction, the initial design of a comprehensive KG schema tailored to specific DES model outputs requires careful upfront domain expertise and engineering effort. Furthermore, although the LLM agent with its self-correction mechanisms performed robustly, the absolute reliability of LLM-generated Cypher queries and the nuanced accuracy of its synthesized explanations warrant ongoing evaluation, particularly when faced with highly novel or ambiguous operational scenarios not extensively represented within the KG's current scope (derived from the simulated data). The generalizability of the specific KG schema and agent fine-tuning has primarily been validated within the described warehouse unloading context, and its seamless applicability to vastly different DES models or a broader array of warehouse processes is yet to be exhaustively demonstrated.

\section{Conclusion and Future Work}
Extracting actionable insights from the complex and voluminous data generated by Discrete Event Simulations poses a significant challenge to timely and effective decision-making in warehouse operations. To address this, we proposed a novel framework that integrates Knowledge Graphs with a reasoning-capable LLM agent, offering a more intuitive and powerful means of interacting with simulation data. The architecture combines a QA chain with step-wise guidance and an iterative reasoning chain equipped with sub-questioning, Cypher query generation, and self-reflection. This enables both high-accuracy responses to operational queries and deeper, evidence-driven investigations into system inefficiencies. 
Experimental evaluations demonstrate this framework's proficiency in accurately answering operational questions and, more significantly, its robust capability in performing iterative, evidence-driven investigations to identify operational bottlenecks within simulated scenarios, surpassing traditional baseline methods.

Looking ahead, several exciting avenues for future research emerge. Firstly, we plan to explore the integration and performance of other advanced reasoning-focused Large Language Model architectures or emerging state-of-the-art alternatives to potentially enhance the agent's diagnostic depth and efficiency. Secondly, to further validate and demonstrate the framework's robustness, we will focus on expanding its application to a wider array of warehouse operations beyond unloading—such as slotting design, order picking, loading, and inventory management—which will inherently involve generating more diverse and complex simulated scenarios tailored to these new contexts. This expansion will also necessitate developing rigorous benchmarking methodologies for its investigative question-answering capabilities to formally quantify performance in bottleneck identification tasks across these varied settings. Such extensions will allow for a thorough assessment of the framework's adaptability and utility across a broader spectrum of logistics challenges, including the potential for analyzing larger-scale supply chain simulations.

\begin{acks}
This work is supported by PhiLabs, Quantiphi Inc. We would like to thank Dr. Dagnachew Birru and Mr. Asif Hasan for their continued support. 
\end{acks}

\newpage
\input{input.bbl}

\newpage
\appendix

\section{Appendix}

\subsection{Warehouse Resources}
The Table \ref{tab:resource_information} highlights the various resource types and their respective ids that are modeled in the simulation.
\begin{table}[!h]
\centering
\caption{The resource ids for the different resource present in the simulated scenario}
\label{tab:resource_information}
\begin{tabular}{c|c} 
\hline
Resource Type & Resource IDs                                                                                                                                                                                                                       \\ 
\hline
Supplier      & \begin{tabular}[c]{@{}c@{}}AuroraFarms, BlackSheepDist, CamelCargo,\\DeltaDrops, EvergreenEdge\end{tabular}                                                                                                                        \\ 
\hline
Worker        & \begin{tabular}[c]{@{}c@{}}BW\_02, BW\_00, BW\_01, BW\_03, BW\_09,\\BW\_10, BW\_11, BW\_08, BW\_05, BW\_07, \\BW\_06, BW\_04\end{tabular}                                                                                          \\ 
\hline
AGV           & \begin{tabular}[c]{@{}c@{}}AGV\_10, AGV\_12, AGV\_11, AGV\_00, \\AGV\_01, AGV\_02, AGV\_03, AGV\_13,\\AGV\_14, AGV\_04, AGV\_15, AGV\_05, \\AGV\_16, AGV\_08, AGV\_17, AGV\_09, \\AGV\_07, AGV\_18, AGV\_19, AGV\_06\end{tabular}  \\ 
\hline
Fork Lift     & FL\_00, FL\_01, FL\_04, FL\_02, FL\_03                                                                                                                                                                                             \\ 
\hline
Storage Block & A, B, C, D                                                                                                                                                                                                                       \\
\hline
\end{tabular}
\end{table}

\subsection{KG Schema:}
\label{appendix:KGSchema}
\textbf{Node Properties:}
\begin{itemize}
    \item \textbf{SUPPLIER:} \\
     \texttt{supplier\_id: STRING, arrival\_time: DATETIME, \\ discharge\_start:DATETIME, discharge\_end: DATETIME}
    
    \item \textbf{WORKER:}\\ \texttt{ worker\_id: STRING}
    \item \textbf{AGV:}\\ \texttt{ agv\_id: STRING}
    \item \textbf{FL (Forklift):}\\ \texttt{ forklift\_id: STRING}
    \item \textbf{STORAGE:}\\ \texttt{ block\_id: STRING}
\end{itemize}

\textbf{Relationship Properties:}
\begin{itemize}
\item \textbf{SUPPLIER\_TO\_WORKER} \\ \texttt{package\_id:STRING, worker\_pick\_up\_start:DATETIME}
\item \textbf{WORKER\_TO\_AGV} \\ \texttt{package\_id: STRING, agv\_arrival:DATETIME,\\ agv\_journey\_start:DATETIME, worker\_pick\_up\_end: DATETIME}
\item \textbf{AGV\_TO\_FL} \\ \texttt{package\_id: STRING, agv\_journey\_end: DATETIME, fl\_placement\_start: DATETIME}
\item \textbf{FL\_TO\_STORAGE} \\ \texttt{package\_id: STRING, fl\_placement\_end: DATETIME}
\end{itemize}

\subsection{Investigative Case Studies - Scenario 3}
\label{scenario_3}
\subsubsection{Scenario 3: Forklift Waiting Times and Connection to Discharge Flow}
\begin{table*}[htbp]
\centering
\caption{Case Study 3: What do the differences in forklift waiting times reveal about the discharge flow?}
\label{tab:case_study_forklift_delay}
\footnotesize 
\begin{flushleft}
\vspace{-0.3cm}
\footnotesize{\textbf{Note:} The LLM Agent formulates its own sequence of questions. Potential human expert iterative questions for a forklift delay investigation might include: (1) KPIs per forklift (tasks, durations, idle times)? (2) Which forklifts show consistently longer task/idle times? (3) Operational time breakdown for suspect forklifts? (4) Queue/wait times for packages per forklift? (5) Are delays linked to specific zones/tasks/times for suspect forklifts? (6) How does a suspect forklift's throughput compare? (7) Any error codes/issues reported? Cypher queries are highly condensed conceptual representations.}
\end{flushleft}

\newcolumntype{L}[1]{>{\raggedright\arraybackslash}p{#1}}
\newcolumntype{T}[1]{>{\raggedright\arraybackslash\ttfamily}p{#1}} 

\begin{tabular}{
    L{0.23\textwidth} 
    L{0.23\textwidth} 
    L{0.23\textwidth} 
    T{0.22\textwidth}  
}
\toprule

\textbf{LLM Agent - Iterative Question} & \textbf{LLM Agent - Plan (Condensed)} & \textbf{KG ResultSet (Condensed)} & \textbf{Cypher Query (Condensed)} \\
\midrule

What is the average forklift waiting time and utilization across all blocks during the discharge flow? & 
Calculate FL waiting (AGV journey\_end - FL placement\_start) & FL utilization (active\_time / total\_span) per FL. \& 
FL\_00: AvgWait ~332.9s, Util 0; \newline FL\_01: AvgWait ~48.9s, Util 0; \newline FL\_04: AvgWait ~36.3s, Util 0; & 
\texttt{MATCH (a:AGV)-[]->(fl:FL)-[]->(st:S) RETURN fl.id, avg(wait\_time), SUM(active\_FL)/total\_FL\_span AS util} \\
\cmidrule(lr){1-4}

What are the average package waiting times at the waiting point for each AGV during the discharge flow, and how do these compare to the global average? & 
Calculate package waiting time for AGV arrival (AGV\_arrival - worker\_pickup\_end) per AGV; compare to global average. & 
Highly variable: e.g., AGV\_10: ~3.6m; AGV\_08: ~10.3m. Global Avg: ~5.5m. (Indicates some AGVs contribute to or experience more package wait). & 
\texttt{MATCH (w)-[wta:WTA]->(a:AGV) RETURN a.id, avg(dur(wta.worker\_end, wta.agv\_arrival)).min UNION ... global\_avg} \\
\cmidrule(lr){1-4}

What is the average time taken by each forklift to move a package from the AGV to the storage bay, and how does this compare to the global average? & 
Calculate FL movement time (FL\_placement\_end - FL\_placement\_start) per FL; compare to global average and show difference. & 
FL\_00: AvgMove ~152.2s (28.9s > global avg ~123.3s). Others (FL\_01, FL\_02, FL\_03, FL\_04) were below global average. & 
\texttt{MATCH (a)-[]->(fl:FL)-[]->(st:S) RETURN fl.id, avg(FL\_move) AS t1; ... global\_avg AS t2 RETURN fl.id, t1, t2, t1-t2} \\
\cmidrule(lr){1-4}

What is the average time taken by each AGV to transport a package from the waiting point to the pickup point, and how does this compare to the global average? & 
Calculate AGV transport time (AGV\_journey\_end - AGV\_journey\_start) per AGV; compare to global average. & 
Variable AGV performance: Some AGVs (e.g., AGV\_07, AGV\_09) faster than ~8.7s global avg; others (e.g., AGV\_00, AGV\_08) slower. & 
\texttt{MATCH (w)-[wta:WTA]->(a)-[atf:ATF]->(fl) RETURN a.id, avg(dur(wta.agv\_start, atf.agv\_end)); ...global\_avg} \\
\midrule 
\multicolumn{4}{L{\dimexpr0.22\textwidth + 0.25\textwidth + 0.25\textwidth + 0.22\textwidth + 6\tabcolsep\relax}}{%
   \textbf{Agent Summary:}  \textbf{Forklift FL\_00} is the primary bottleneck within the discharge flow, evidenced by its highest average waiting time (approximately 333 seconds), substantially longer than other forklifts, and also by it taking significantly more time to move packages from AGVs to storage (around 152 seconds, which is about 29 seconds above the global average for this task). While variability was also observed in AGV performance and in total unload times among different suppliers, addressing the specific operational challenges and inefficiencies tied to FL\_00 can help improve the overall efficiency of the discharge process.
} \\
\bottomrule
\end{tabular}
\end{table*}

In the third investigative case study, which implicitly probed for forklift-induced delays, the human expert insightfully identified Forklift FL\_00 (and an associated block) as being linked to above-average AGV differences or wait times, though with some initial uncertainty about whether AGVs, workers, or forklifts were solely responsible. The baseline Direct QA + Self-Reflection method also flagged FL\_00 for having the highest average waiting time (~333s) but offered limited further diagnosis beyond noting variations among forklifts. Our method (Table \ref{tab:case_study_forklift_delay}), through its sequence of targeted inquiries, not only strongly corroborated FL\_00's problematic status but also provided a more comprehensive diagnostic picture. The agent confirmed FL\_00's exceptionally high average waiting time and, crucially, also identified its significantly longer package movement time from AGV to storage (~152s vs. ~123s global average). While also exploring AGV performance variability, the agent's overall diagnosis clearly centered on FL\_00's dual inefficiency (high waiting and long task execution) as a key contributor to delays, thus offering a more multi-faceted and data-rich explanation than the baseline and a more conclusive identification of forklift-specific issues compared to the human expert's initial broader considerations.

\subsection{Operational QA - Success and Failure Cases}
\label{operational_qa_success_failure}
We present selected query instances from Operational QA to qualitatively compare our proposed method against baseline approaches. These examples highlight specific successes (marked in \textcolor{darkgreen}{green}) and failures (marked in \textcolor{red}{red}), offering deeper insight into the practical performance of each technique.

\subsection*{Case 1: Baseline (Direct QA) Partial Failure, Other Methods Pass}
\noindent\textbf{Question:} Which supplier had the shortest total discharge time and how many packages were moved?

\medskip 

\noindent\textbf{Direct QA:} The supplier "EvergreenEdge" had the shortest total discharge time, which was 3562 seconds. \textcolor{red}{A total of 1 package was moved.}

\medskip

\noindent\textbf{Direct QA + SR:} \textcolor{darkgreen}{The supplier with the shortest total discharge time is EvergreenEdge, and they moved 33 packages.}

\medskip

\noindent\textbf{Step-wise Guide:} \textcolor{darkgreen}{The supplier with the shortest total discharge time is EvergreenEdge, with a discharge time of 3562 seconds. They moved a total of 33 packages.}

\medskip

In this instance, the query required retrieving two distinct pieces of information (the supplier with the shortest discharge time and their corresponding package count) and ensuring their correct association. The `Direct QA' baseline successfully identified the correct supplier and their discharge time but faltered in accurately retrieving the associated package count, incorrectly stating \textit{1 package}. This likely indicates an inability of the single-pass query generation to correctly formulate or execute a Cypher query complex enough to link these two facts or perform the necessary aggregation accurately. The introduction of self-reflection in the `Direct QA + SR' baseline enabled it to overcome this error, suggesting the reflection process could validate and correct the factual components of the synthesized answer. Our `Step-wise Guide' approach also provided the correct and complete answer. Its structured approach, which can decompose the query into logical steps (e.g., one for identifying the supplier with the shortest time, followed by a step to retrieve the package count specifically for that identified supplier) and validate each part, inherently leads to higher accuracy for such multi-fact retrieval questions.

\subsection*{Case 2: Both Baselines Fail, Step-wise Guide Succeeds}

\noindent\textbf{Question:} What is the average travel time for an AGV to move a package from the dock to its assigned storage area?

\medskip

\noindent\textbf{Direct QA:} \textcolor{red}{The average travel time for an AGV to move a package from the dock to its assigned storage area is approximately 178 seconds.}

\medskip

\noindent\textbf{Direct QA + SR:} \textcolor{red}{The average travel time for an AGV to move a package from the dock to its assigned storage area is approximately 178.14 seconds.}

\medskip

\noindent\textbf{Step-wise Guide:} \textcolor{darkgreen}{The average travel time for an AGV to move a package from the dock to its assigned storage area is 455 seconds.}

\medskip

This case highlights a scenario where both baseline methods failed to determine the correct numerical value for a key performance indicator. Their answers (178s and 178.14s) were significantly different from the correct value (455s) obtained by the Step-wise Guide. This discrepancy suggests that the baselines likely misinterpreted the scope of the AGV travel (e.g., focusing on a shorter segment of the journey, incorrectly identifying the start/end events for the "dock to storage" path, or using an incorrect aggregation method over the KG). The Direct QA + SR baseline, while providing a slight numerical variation, essentially repeated the same fundamental error as the Direct QA, indicating that its self-reflection mechanism was insufficient to catch this type of error in calculation logic or path identification. The success of the proposed method implies that its structured decomposition of the query into smaller, verifiable steps allowed it to accurately identify the relevant AGV journey segments (from dock to storage for each package) and correctly aggregate these travel times from the KG to compute the true average.

\subsection*{Case 3: Step-wise Guide Fails, Baseline (QA + SR) Succeeds}

\noindent\textbf{Question:} For each person, what was the total number of packages they handled during a shift?

\medskip

\noindent\textbf{Direct QA:} \textcolor{red}{Syntax Error}

\medskip

\noindent\textbf{Direct QA + SR:} \textcolor{darkgreen}{Worker BW\_11 handled 13 packages, Worker BW\_10 handled 19 packages, Worker BW\_09 handled 20 packages, Worker BW\_12 handled 18 packages, Worker BW\_05 handled 9 packages, Worker BW\_08 handled 8 packages, Worker BW\_06 handled 9 packages, Worker BW\_07 handled 9 packages, Worker BW\_02 handled 13 packages, Worker BW\_00 handled 14 packages, Worker BW\_01 handled 14 packages, Worker BW\_03 handled 15 packages, and Worker BW\_04 handled 12 packages.}

\medskip

\noindent\textbf{Step-wise Guide Response:} \textcolor{red}{The context does not provide any information about the workers or the number of packages they handled during a shift. Therefore, it is not possible to determine the total number of packages handled by each worker.}

\medskip

This case presents an interesting scenario where the Step-wise Guide incorrectly concluded that the information was unavailable, while the Direct QA + SR baseline successfully retrieved the correct data. The Direct QA baseline failed completely with a syntax error, likely struggling to generate the somewhat complex Cypher query required for a group-wise aggregation (summing packages per worker). The Direct QA + SR, however, managed to overcome this, indicating that its combined query generation and self-reflection capability was sufficient for this particular aggregation task. The failure of the Step-wise Guide ("context does not provide information") suggests a potential limitation in its current decomposition strategy or schema interpretation when faced with "for each" type queries requiring specific group-by-and-aggregate operations. It's possible that its step-generation logic broke down the problem in a way that obscured the path to aggregation, or it failed to correctly map person to the Worker entity and their associated package handling events in a way that allowed for summation. This highlights an area for future refinement in the step generation and KG traversal logic within our proposed method to better handle such complex aggregation queries.

\begin{table*}[ht]
\centering
\caption{Set of 25 Operational Questions along with their categorization.}
\label{tab:operational_questions}
\begin{tabular}{|l|p{14cm}|} 
\hline
\textbf{Category} & \textbf{Question} \\ 
\hline
SUPPLIER & What is the number of discharge processes that are completed on an hourly basis? 
\\ 
 & Where and how many containers discharged from supplier DeltaDrops distributed in each block in the storage?              \\ 
 & Which supplier had the shortest total discharge time and how many packages were moved?                                   \\ 
 & What is the average waiting time for a supplier truck before unloading begins? Which truck waited the most?              \\ 
 & Which hour had the most total waiting time during package unload?                                                       \\ 
\hline

WORKER   & For each person, what was the total number of packages they handled during a shift?                                      \\ 
   & What is the average time taken by a person to move a package from truck to  AGV? Who is the most efficient person ?  \\ 
   & How much time does each worker take to unload all packages from supplier DeltaDrops?                                     \\ 
   & How many workers were used to unload packages from supplier CamelCargo?                                     \\ 
   & Which workers were assigned to most number of suppliers?     \\                    \hline                            
AGV      & Which three AGVs processed the least amount of packages?                                                                 \\ 
      & What is the average travel time for an AGV to move a package from the dock to its assigned storage area?                 \\ 
      & How many trips does each agv make during unloading along with the average journey time?                                  \\ 
& How many packages did AGV 04 handle from each supplier ? \\
& Which AGV was the least utilized ?\\

\hline

FORKLIFT & Which package waited the longest for a fork lift ?                                                                       \\ 
 & How many packages are handled by each forklift?                                                                          \\ 
 & Which forklift is the most under utilized ?                                                                              \\ 
 & What is the average time taken by a forklift to move a package to its assigned storage space?                            \\ 
 & What is the utilization rate (percentage of time in use) for each forklift?                                              \\ 
\hline
PACKAGE  & which storage block contains the highest number of containers?                                                           \\ 
  & What is the average time a package discharge takes?                                                                      \\ 
  & What is the average waiting time for a package to be transferred to a forklift after AGV arrival at the storage area?    \\ 
  & Which package experienced the longest total time from arrival at the dock to placement in its final storage location?    \\ 

  & How many packages took longer than the average unload time during and what is the average discharge time?                \\ 
  & Which packages were handled by both agv 10 and forklift 00?                                                              \\ 
\hline
\end{tabular}
\end{table*}

\end{document}